\documentclass{article}
\usepackage{nips15submit_e,times}
\usepackage{graphicx}
\usepackage{hyperref}
\usepackage{url}

\title{Detecting bird sound in unknown acoustic background using crowdsourced training data}

\author{
Timos Papadopoulos \\
Long Term Ecology Lab and \\
Machine Learning Group \\
University of Oxford \\
\texttt{timosp@zoo.ox.ac.uk} \\
\And
Stephen Roberts \\
Machine Learning Group \\
University of Oxford \\
\texttt{sjrob@robots.ox.ac.uk} \\
\And
Kathy Willis \\
Long Term Ecology Lab \\
University of Oxford \\
\texttt{kathy.willis@zoo.ox.ac.uk} \\
}

\nipsfinalcopy 

\begin{document}

\maketitle

\begin{abstract}
Biodiversity monitoring using audio recordings is achievable at a truly global scale via large-scale deployment of inexpensive, unattended recording stations or by large-scale crowdsourcing using recording and species recognition on mobile devices. The ability, however, to reliably identify vocalising animal species is limited by the fact that acoustic signatures of interest in such recordings are typically embedded in a diverse and complex acoustic background. To avoid the problems associated with modelling such backgrounds, we build generative models of bird sounds and use the concept of novelty detection to screen recordings to detect sections of data which are likely bird vocalisations. We present detection results against various acoustic environments and different signal-to-noise ratios. We discuss the issues related to selecting the cost function and setting detection thresholds in such algorithms. Our methods are designed to be scalable and automatically applicable to arbitrary selections of species depending on the specific geographic region and time period of deployment.
\end{abstract}

\section{Introduction}

The present day availability of cheap recording devices, widespread use of mobile devices and availability of network connectivity make possible the recording, storage and online accessing of environmental sound at an unprecedented level. In order for this large collection of data to be useful for the purposes of biodiversity monitoring, methods are needed for the automatic identification of vocalising animals. In the case of bird sound identification that is considered here, most previous work is focused on the problem of classifying a given sound interval that is known to predominantly contain bird sound among a predetermined set of classes (usually corresponding to different bird species). However, such work is not directly applicable to recordings obtained in the way described above. This is because such recordings are typically sparse in bird sound but can also contain any number of other sources of sound. The output of a classifier trained and optimised to discriminate between different classes of bird sound can be problematic to interpret and evaluate for such non-bird sound inputs. At the same time, designing, training and testing in a statistically balanced way, classifiers that include a model for any type of possible acoustic background is not a practical option.

An alternative, hierarchical, approach would be to include a screening stage where bird sound is first detected against other sources and then classified to species classes. In this work we take the approach of building a probabilistic generative normality model  for each one of the classes of interest (different bird species). Our objective is for each one of these individual models to be adequately discriminative against non-bird sound; discrimination between different classes of bird sound is not the primary target. Designing this first stage in a principled probabilistic manner facilitates the effective and versatile integration of its inference and decision output into a subsequent classification stage. That classification stage can be flexible in terms of the bird species (classes) considered, for example by using a `lazy' classifier where the classes to be considered are chosen after the training stage. Further to that, building per-species normality models can be directly used in human-machine cooperation schemes where, for instance, very large collections of recorded data are drastically condensed to a probabilistically ordered list of inferred occurrences of any chosen class of interest and presented to a human expert for validation.

Previous work on birdsong detection (\cite{anderson96,kogan98,chen06,chu11}) rely on the existence of a cleanly annotated dataset for the training of the automatic recognition system; in many cases segmented and annotated down to the syllable level. The creation of such cleanly annotated training data is laborious and cannot feasibly be extended to cover the total number of (approximately 10000) bird species. In addition to that, results obtained on such fragmented or proprietary training and testing data are not easy to reproduce independently and evaluate in comparison with other methods. On the other hand, audio recordings of virtually world-wide bird species distribution, albeit with less detailed annotation metadata, are currently openly available on crowdsourced platforms. A prime example of that is the \emph{xeno-canto} website (\url{www.xeno-canto.org}) which currently provides in excess of 200000 recordings of more than 9000 avian species. We make use of the less detailed annotation that is provided by individual users in that collection to obtain training data for the determination of generative models for 15 species. We subsequently test the effectiveness of these models in determining the location of bird sound that we have added at random positions inside background sound recordings.

The objective of the work presented in this paper is to examine the effectiveness and performance limitations of a baseline audio feature extraction and normality model estimation method and to identify directions for the improvement of its performance. The methods we present are completely free of any manual data preprocessing and hence (in conjunction with the aforementioned source of available data) directly scalable to larger numbers of bird species. 

\section{Methods}

The data we use for testing our detection method consist of environmental sounds from the `IEEE AASP Challenge: Detection and Classification of Acoustic Scenes and Events competition database'\footnote{\url{http://c4dm.eecs.qmul.ac.uk/sceneseventschallenge}}. Separate rounds of experiments were run with randomly selected recordings from the `Park' and the `Open air market' categories. The former of these categories is chosen as representative of an urban natural environment; it contains recordings with larger segments of silence and it is expected to be less confounding for the detection task. Recordings from the latter category are very dense in human speech and busy urban environment sounds and are expected to be more challenging for the detection task. The clean bird sound data are recordings of various time lengths (ranging from approximately 0.5s to 10s) obtained form the `Reference Animal Vocalisations' database of the `Animal Sound Archive of the Museum für Naturkunde Berlin'\footnote{\url{http://www.animalsoundarchive.org/RefSys/Preview.php}}. We choose 15 bird species for which that database has more than 100 recording samples (however we only use recordings that are `Open Access'). In the results presented in the next section, species numbers are as the alphabetical ordering in Table~\ref{tab:specfeat}.

\begin{table}[t]
	\caption{List of species, features and feature vectors used for the design of the bird sound normality models in different experimental cases.}
	\label{tab:specfeat}
	\vspace{\baselineskip}
	\centerline{
	\begin{tabular}{ c | c  || c | c }
		   & \textbf{Species}				&	&		\textbf{Features} \\
		\hline
  		1 & Emberiza hortulana 		& 1 	& 		Mean \\
  		2 & Emberiza schoeniclus 	& 2 	& 		Standard Deviation \\
  		3 & Fringilla coelebs 			& 3 	& 		Skewness \\
  		4 & Luscinia luscinia 			& 4 	& 		Kurtosis \\
  		5 & Luscinia megarhynchos 	& 5 	& 		Mode \\
  		6 & Parus major 			& 6 	& 		SFM \\
  		7 & Periparus ater 			&  	&  		\textbf{Feature sets examined} \\
  		8 & Phoenicurus phoenicurus 	&  	& [1] \\
  		9 & Phylloscopus collybita 	& 	& [5] \\
  		10 & Phylloscopus ibericus 	& 	& [1 2] \\
  		11 & Phylloscopus trochilus 	& 	& [5 2] \\
  		12 & Rallus aquaticus 		& 	& [1 2 6] \\
  		13 & Sylvia atricapilla 		& 	& [5 2 6] \\
  		14 & Turdus merula 			& 	& [1 2 3 4] \\
  		15 & Turdus philomelos 		& 	& [1 2 3 4 5 6] \\
	\end{tabular}
	}
\end{table}

For the training data we use recordings from the \emph{xeno-canto} database labelled with the same 15 species chosen for testing. For each species we select only recordings labelled with an `A' (highest) quality rating and that are annotated as not having other bird species in the background. We compute spectrograms for each recording (20ms frame length, 50\% overlap, FFT length equal to the closest power of two that gives at least 93Hz frequency bin spacing, for the sampling rate of the specific recording). We only keep frequency bins in the range of 1-10kHz. This frequency range retains the vocalisation frequency content of most avian species while discarding low frequency wind noise and recorder self noise and it is typical in the related literature. The recordings in \emph{xeno-canto} typically contain long intervals of other ambient or anthropogenic sounds, but in recordings with an `A' quality rating these non-bird intervals are predominantly silence or low frequency wind or recorder self-noise. In each recording's spectrogram we discard frames with total power less than the 90th percentile expecting that the selected 10\% more energetic frames come nearly exclusively from bird sound time intervals (we selectively verified that by listening). Each of the selected spectrogram frames is normalised to unity sum (hence having the form of a probability mass function, pmf) and 6 spectral statistics listed in Table~\ref{tab:specfeat} are extracted as features. Quadratic interpolation between two frequency bins surrounding the maximum is applied for the determination of the frequency position of maximum frequency (mode of the pmf). The Spectral Flatness Measure is computed with the method described in \cite{madhu09}. Each element in the 6-element feature vector is standardised to zero mean and unit standard deviation via a simple (invertible) linear transform.

For each species we randomly select 6000 feature vectors (given the spectrogram computation overlap and the 90th percentile selection, this corresponds to 60-120sec of `bird only' frames or equivalently to 10-20min actual recording duration) and we use it to train a full covariance Gaussian Mixture Model (GMM) using Expectation Maximisation (EM) (for a detailed description of GMM fitting using EM see \cite{bishop06} ch.9). GMMs are fitted to different selections of feature sets ranging from univariate up to the full 6 dimensional feature vector as listed in Table~\ref{tab:specfeat}. In each case the EM algorithm is run 10 times for different random initialisations and the GMM converged to the highest log-likelihood is selected. We set the number of components to vary from 1 to 15. The Minimum Description Length (MDL, \cite{oliver96}) model selection criterion is computed for each one of the 1- to 15-component GMMs.

\begin{figure}[t]
    \centering
    \includegraphics[width=\textwidth]{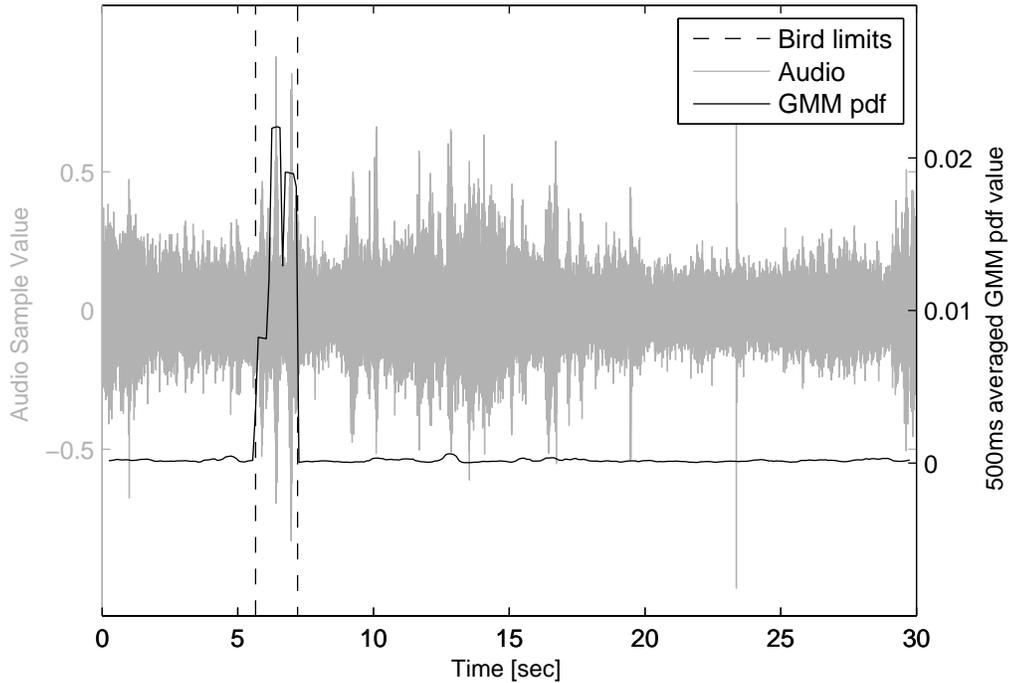}
    \caption{Example: Audio signal (grey line) with marked bird limits (black dashed line) and value of the corresponding species GMM pdf averaged over a 500ms interval.}
    \label{fig:example1}
\end{figure}

The same feature extraction process is applied to the test signals, but keeping all the frames (no 10\% power thresholding). The computed test features are normalised with the mean and standard deviation values computed from the training feature set. The value of the probability density function (pdf) for each test frame feature vector is obtained from the corresponding species' GMM  and per-frame pdf values are averaged over consecutive frames spanning a 500ms time interval. Averaged GMM pdf values are taken at 100ms steps. The Receiver Operating Characteristic curve (ROC curve) is computed based on whether the 500ms interval (or the larger part of it) is within the limits of the randomly chosen placement of the bird signal. For a binary classification method that assigns a probability (or probabilistically interpretable score) of class membership to each given test instance, the ROC Curve traces the True Positive Rate versus False Positive Rate points obtained when the threshold of class separation moves continuously from minus infinity to plus infinity, thus effectively evaluating the performance of the classifier in correctly sorting the given instances (in our case 500ms intervals) with respect to membership in the two classes (in our cases time intervals known to containing bird sound compared to time intervals of background sound). The corresponding Area Under the Curve (AUC) metric is one effective way to evaluate such classifier discrimination performance in cases of imbalanced proportion of positive to negative examples in the test set (for the related details see \cite{fawcett06}). The results of the next section are median values of the AUC obtained over 50 repetitions of test signals as described above. Figure~\ref{fig:example1} provides an example of such a test sample.

\section{Results}

\begin{figure}[ht]
    \centering
    \includegraphics[width=\textwidth]{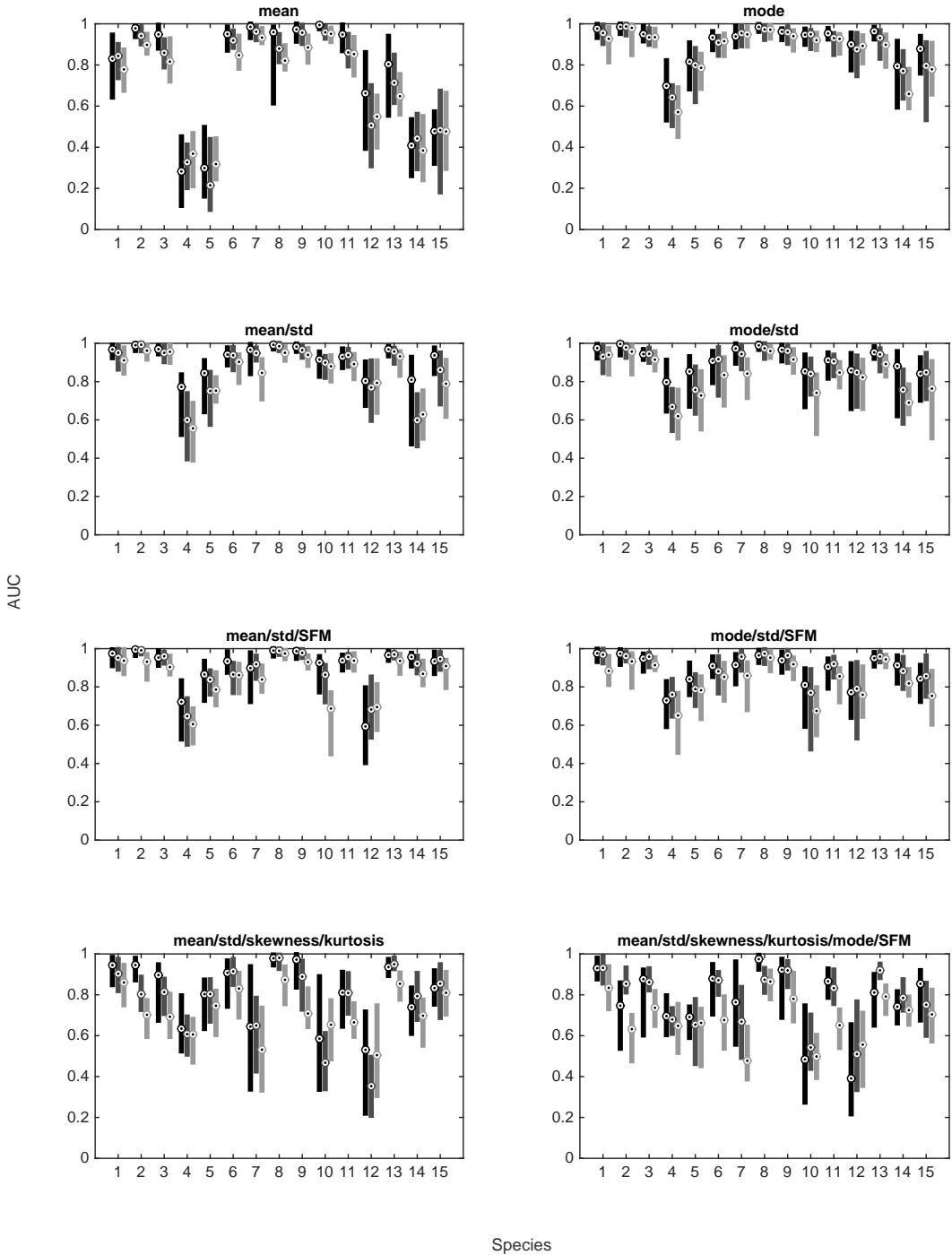}
    \caption{Median (dotted circle) and interquartile range (25th to 75th percentile - vertical bar) of the AUC over 50 tests. Three cases of background (`Park' category with +3dB SNR, `Open air market' category with +3dB and -3dB SNR) are plotted with a horizontal shift and with black, grey and light grey colours respectively.}
    \label{fig:fig1_collect}
\end{figure}

Figure~\ref{fig:fig1_collect} summarily presents the per-species detection performance (median and interquartile range of the AUC over 50 tests) obtained when a bird sound normality model fitted to each individual species' training data is used to detect vocalisations of that species embedded at random positions (different in each individual test) in a non-bird background. The results obtained with GMM density estimation models for each of the 8 feature combinations listed in Table~\ref{tab:specfeat} are plotted in separate subplots. Three different cases of background are considered, namely recordings from the `Park' category with the SNR set to +3dB and from the `Open air market' category with the SNR set to +3dB and -3dB. We note that each set of results for the three different backgrounds was obtained with a different draw of 6000 frames for the GMM modelling.The results plotted in Figure~\ref{fig:fig1_collect} are those obtained with the GMM with a number of components as determined by the minimum value of the MDL criterion over the range of 1 to 15 components. The same results are also summarised in a tabular form in Table~\ref{tab:results}. 

\begin{table}[t]\scriptsize
	\caption{Median of the AUC over 50 tests. Columns correspond to feature vectors and rows to species, both in the same order as listed in Table~\ref{tab:specfeat}. The first line of each cell is the median AUC and the second line is the number of components as chosen by the minimum of the MDL criterion for the three cases of background considered.}
	\label{tab:results}
	\vspace{\baselineskip}
\centerline{
	\begin{tabular}{|c | c | c | c | c | c | c | c|}
		\hline
                    0.83-0.84-0.78 &0.98-0.96-0.93 &0.97-0.95-0.91 &0.98-0.93-0.94 &0.98-0.95-0.94 &0.98-0.97-0.88 &0.95-0.90-0.86 &0.93-0.93-0.83\\
                    3-4-3 &6-9-6 &5-6-6 &15-15-8 &10-8-11 &15-12-9 &15-15-15 &15-15-15\\
		\hline
                    0.98-0.94-0.90 &0.99-0.99-0.98 &0.99-0.99-0.96 &1.00-0.98-0.96 &1.00-0.99-0.93 &0.97-0.96-0.93 &0.95-0.80-0.70 &0.75-0.85-0.63\\
                    4-4-3 &5-4-8 &6-6-5 &8-6-7 &9-10-7 &9-8-10 &15-15-15 &15-15-15\\
		\hline
                    0.95-0.86-0.82 &0.95-0.93-0.93 &0.97-0.95-0.96 &0.94-0.95-0.91 &0.95-0.96-0.90 &0.95-0.96-0.91 &0.90-0.81-0.69 &0.88-0.86-0.74\\
                    3-3-2 &5-4-4 &5-5-5 &6-8-5 &8-9-7 &8-10-11 &15-15-15 &15-15-15\\
		\hline
                    0.28-0.33-0.37 &0.70-0.64-0.57 &0.77-0.60-0.56 &0.80-0.67-0.62 &0.72-0.65-0.61 &0.73-0.76-0.65 &0.63-0.61-0.61 &0.70-0.68-0.65\\
                    3-3-3 &5-5-6 &6-6-7 &9-7-7 &10-12-11 &11-10-10 &15-15-15 &15-15-15\\
		\hline
                    0.30-0.21-0.32 &0.82-0.80-0.79 &0.84-0.75-0.75 &0.85-0.76-0.73 &0.87-0.84-0.79 &0.84-0.79-0.78 &0.80-0.80-0.75 &0.69-0.65-0.66\\
                    3-3-3 &5-5-7 &6-5-4 &7-6-5 &10-7-8 &8-7-9 &15-15-15 &15-15-15\\
		\hline
                    0.95-0.92-0.85 &0.93-0.91-0.92 &0.94-0.94-0.90 &0.91-0.92-0.83 &0.93-0.87-0.86 &0.91-0.88-0.85 &0.91-0.91-0.83 &0.88-0.87-0.68\\
                    3-3-3 &5-4-4 &5-5-6 &6-7-6 &10-9-11 &11-10-12 &15-15-15 &15-15-15\\
		\hline
                    0.98-0.96-0.93 &0.94-0.95-0.95 &0.97-0.95-0.85 &0.97-0.94-0.84 &0.90-0.92-0.84 &0.91-0.96-0.86 &0.64-0.65-0.53 &0.76-0.67-0.48\\
                    3-3-2 &4-4-4 &8-9-10 &10-9-11 &12-9-12 &13-13-9 &15-15-15 &15-15-15\\
		\hline
                    0.96-0.88-0.82 &0.98-0.97-0.97 &0.99-0.98-0.95 &0.99-0.97-0.96 &0.99-0.99-0.97 &0.96-0.97-0.95 &0.98-0.98-0.87 &0.97-0.87-0.86\\
                    3-4-4 &5-3-3 &7-6-8 &6-6-8 &9-7-8 &9-9-8 &15-15-15 &15-15-15\\
		\hline
                    0.97-0.96-0.88 &0.96-0.96-0.94 &0.98-0.96-0.94 &0.97-0.96-0.92 &0.99-0.98-0.93 &0.94-0.96-0.92 &0.97-0.89-0.71 &0.92-0.92-0.78\\
                    3-3-3 &6-4-10 &6-6-7 &7-9-7 &9-9-8 &9-8-11 &15-15-15 &15-15-15\\
		\hline
                    0.99-0.96-0.95 &0.95-0.95-0.92 &0.92-0.90-0.88 &0.85-0.84-0.74 &0.93-0.86-0.69 &0.81-0.77-0.67 &0.59-0.47-0.65 &0.48-0.54-0.50\\
                    3-2-5 &5-5-8 &6-7-5 &8-10-10 &12-12-10 &13-12-11 &14-15-15 &15-15-15\\
		\hline
                    0.95-0.86-0.85 &0.95-0.93-0.93 &0.93-0.94-0.89 &0.91-0.91-0.85 &0.94-0.96-0.94 &0.91-0.92-0.86 &0.81-0.81-0.67 &0.87-0.83-0.65\\
                    3-4-3 &5-4-6 &11-9-9 &8-8-9 &9-10-10 &8-8-10 &15-15-15 &15-15-15\\
		\hline
                    0.66-0.51-0.55 &0.90-0.88-0.89 &0.80-0.77-0.79 &0.86-0.85-0.82 &0.59-0.68-0.69 &0.77-0.79-0.76 &0.53-0.35-0.50 &0.39-0.51-0.56\\
                    3-3-3 &6-7-10 &8-7-7 &7-10-8 &10-9-11 &9-12-9 &15-15-15 &15-15-15\\
		\hline
                    0.80-0.71-0.65 &0.96-0.93-0.90 &0.97-0.96-0.93 &0.95-0.94-0.89 &0.97-0.97-0.94 &0.95-0.96-0.94 &0.94-0.95-0.85 &0.81-0.92-0.79\\
                    3-3-3 &3-3-3 &3-3-3 &5-5-4 &6-8-6 &7-7-7 &15-15-15 &15-15-15\\
		\hline
                    0.41-0.44-0.38 &0.79-0.77-0.66 &0.81-0.60-0.63 &0.88-0.76-0.69 &0.96-0.92-0.87 &0.91-0.88-0.82 &0.74-0.79-0.70 &0.74-0.78-0.73\\
                    4-4-4 &5-5-5 &6-7-7 &7-7-7 &10-9-8 &12-12-11 &15-15-15 &15-15-15\\
		\hline
                    0.48-0.48-0.48 &0.88-0.80-0.78 &0.94-0.86-0.79 &0.84-0.85-0.76 &0.93-0.95-0.91 &0.84-0.86-0.75 &0.83-0.85-0.81 &0.85-0.75-0.70\\
                    4-3-3 &3-8-5 &4-4-4 &7-7-8 &8-7-8 &9-7-8 &15-15-15 &15-15-15\\
		\hline
	\end{tabular}
}
\end{table}

As can be seen in Figure~\ref{fig:fig1_collect}, the best results overall are obtained by using only one feature, namely the mode, with a univariate GMM. In that case the median AUC obtained by the MDL-selected GMM is higher than 0.9 for 11 of the 15 species in `Park' +3dB SNR background case and for 10 of the 15 species in the other background cases. The lowest median AUC is 0.70, 0.64 and 0.56 for the three background cases (all for the \emph{Luscinia luscinia} species). The discriminative power of this particular feature is not surprising considering previously published classification results that rely on frequency peak tracking based features (\cite{fager05,somervuo06}). On the other hand, overall, the higher dimension models do not achieve to enrich the normality model (bird sound) so as to more successfully discriminate against background. For particular species, (\emph{Turdus merula} and \emph{Turdus philomelos}) the inclusion of the SFM feature in the model appears to achieve consistently better results. Even though the detection method investigated here can, in principle, be applied with normality models based on different selections of features (possibly individually optimised for each different species), a more detailed examination of such a possibility is required to support the practical merit of such an option.

Comparison of the results obtained with the three different background cases shows an expected but limited degradation between the `Park' and `Open air market' cases and, in turn, when the SNR is reduced by 6dB in the latter case. The consistency of these comparative results is encouraging when taking into account that they are obtained with GMM density estimates trained on different draws of frames from the training data pool and different random selections of bird/background tests. The number of GMM components determined by the MDL criterion (see even numbered rows of Table~\ref{tab:results}) is also generally consistent across different species and feature vector cases. In the cases of the 4- and 6- dimensional models considered here (last two columns of Table~\ref{tab:results}) the MDL minimum is not achieved within the 1 to 15 components range but (as we have verified in separate experiments) the performance does not improve consistently when the number of components is increased up to and beyond the optimum MDL point.

\section{Discussion - Conclusion}

The results presented above provide support for the main objective of this piece of research. We make use of a source of data (\emph{xeno-canto}) that covers a very large and constantly increasing proportion of the total number of vocalising avian species but provides less detailed labelling than is typically used in most published works on the subject. We train detection models for individual species in a way that does not involve any manual preprocessing and is directly scalable to any selection of species. The normality models we learn from these data are shown to indeed achieve discrimination against non-bird sound background which we make no effort to model. Repetitions of the training algorithm over different randomly selected frame collections yield consistent detection results.

On the other hand, the baseline approach presented here underperforms for a number of the species (namely Luscinia luscinia, Luscinia megarhynchos and Turdus merula among the species considered here). In the case of the best performing `mode' scalar feature vector, the density estimates for these species (not presented here) have their main lobe lower in frequency (in the region around 1-2kHz) and are thus closer to the characteristics of urban sounds and human speech. In recordings where the acoustic background is more dense in other sources of biophony (non-bird aminal vocalisations) we would expect the discriminative ability of other species' density models which are distributed higher in the frequency mode feature dimension to also deteriorate.

The choice to build a normality model based on features obtained from statistics of spectrogram frames (rather than, e.g. Mel-Frequency Cepstral Coefficient (MFCC) based features; another widely employed choice) was partly motivated by positive results previously presented in similar recognition tasks by the use of such features (see e.g. \cite{briggs12,fager05,somervuo06}) and equally with the aim of keeping the feature vector dimension low in the interest of better convergence and interpretability properties of the GMM fitting. However, a basic premise of the approach investigated here, namely that a richer (and hence more discriminative) normality model can be obtained by using higher dimension feature vectors is not justified by our current results. A direction that we are currently investigating aiming to address this shortcoming is the incorporation of time-dependence modelling in the learning scheme by fitting a Hidden Markov Model (HMM) to training data obtained in the same way as described above.

Finally it must be noted that, while the focus of this work is to evaluate the discriminative power of this specific detection method, the practical use of such a detection method also requires the application of a threshold level determining the decision boundary between normal and abnormal test inputs (\cite{pimentel14}). A principled probabilistic method for the determination of such a boundary for the specific type of Gaussian mixture density models which based on Extreme Value Theory is described in \cite{roberts00}. We are currently investigating the practical effectiveness of threshold determination methods for the specific bioacoustic detection problem at hand.

\bibliography{papadopoulos_etal}

\begin{thebibliography}{10}

\bibitem{anderson96}
S.~E. Anderson, A.~S. Dave, and D.~Margoliash.
\newblock Template-based automatic recognition of birdsong syllables from
  continuous recordings.
\newblock {\em Journal of the Acoustical Society of America},
  100(2):1209--1219, 1996.

\bibitem{kogan98}
J.~A. Kogan and D.~Margoliash.
\newblock Automated recognition of bird song elements from continuous
  recordings using dynamic time warping and hidden markov models: A comparative
  study.
\newblock {\em Journal of the Acoustical Society of America},
  103(4):2185--2196, 1998.

\bibitem{chen06}
Zhixin Chen and Robert~C. Maher.
\newblock Semi-automatic classification of bird vocalizations using spectral
  peak tracks.
\newblock {\em The Journal of the Acoustical Society of America},
  120(5):2974--2984, 2006.

\bibitem{chu11}
Chu Wei and D.~T. Blumstein.
\newblock Noise robust bird song detection using syllable pattern-based hidden
  markov models.
\newblock In {\em IEEE International Conference on Acoustics, Speech and Signal
  Processing (ICASSP)}, pages 345--348, 2011.

\bibitem{madhu09}
N.~Madhu.
\newblock Note on measures for spectral flatness.
\newblock {\em Electronics Letters}, 45(23):1195, 2009.

\bibitem{bishop06}
Christopher~M Bishop.
\newblock {\em Pattern recognition and machine learning}.
\newblock Springer, New York, 2006.

\bibitem{oliver96}
Jonathan~J Oliver, Rohan~a Baxter, and Chris~S Wallace.
\newblock Unsupervised learning using mml.
\newblock In {\em Proc. 13th Int. Conf. Machine Learning (ICML96)}, pages
  364--372, San Francisco, 1996.

\bibitem{fawcett06}
Tom Fawcett.
\newblock {An introduction to ROC analysis}.
\newblock {\em Pattern Recognition Letters}, 27(8):861--874, 2006.

\bibitem{fager05}
S.~Fagerlund and A.~Harma.
\newblock Parametrization of inharmonic bird sounds for automatic recognition.
\newblock In {\em 13th European Signal Processing Conference (EUSIPCO ’05)},
  2005.

\bibitem{somervuo06}
P.~Somervuo, A.~Harma, and S.~Fagerlund.
\newblock Parametric representations of bird sounds for automatic species
  recognition.
\newblock {\em IEEE Transactions on Audio, Speech, and Language Processing},
  14(6):2252--2263, 2006.

\bibitem{briggs12}
F.~Briggs, B.~Lakshminarayanan, L.~Neal, X.~Z. Fern, R.~Raich, S.~J.~K. Hadley,
  A.~S. Hadley, and M.~G. Betts.
\newblock Acoustic classification of multiple simultaneous bird species: A
  multi-instance multi-label approach.
\newblock {\em Journal of the Acoustical Society of America},
  131(6):4640--4650, 2012.

\bibitem{pimentel14}
Marco~A.F. Pimentel, David~A. Clifton, Lei Clifton, and Lionel Tarassenko.
\newblock {A review of novelty detection}.
\newblock {\em Signal Processing}, 99:215--249, June 2014.

\bibitem{roberts00}
S.J. Roberts.
\newblock {Extreme value statistics for novelty detection in biomedical data
  processing}.
\newblock {\em IEE Proceedings - Science, Measurement and Technology},
  147(6):363--367, November 2000.

\end{thebibliography}
\bibliographystyle{unsrt}

\end{document}